\documentclass[a4paper,conference]{IEEEtran}
\ifCLASSINFOpdf
\usepackage{amsmath}
\usepackage{siunitx}
\usepackage[pdftex]{graphicx}
\usepackage{multicol}
\else
\fi
\begin{document}
%
\title{\huge Tactile-ViewGCN: Learning Shape Descriptor from Tactile Data using Graph Convolutional Network}

\author{
\IEEEauthorblockN{Sachidanand VS}
\IEEEauthorblockA{Department of Physics,\\
IIT Madras, 600036, India\\
ep19b010@smail.iitm.ac.in
}
\and
\IEEEauthorblockN{Mansi Sharma}
\IEEEauthorblockA{Department of Electrical Engineering,\\
IIT Madras, 600036, India\\
mansisharmaiitd@gmail.com, mansisharma@ee.iitm.ac.in
}
}

\maketitle

\begin{abstract}
For humans, our ``senses of touch'' have always been necessary for our ability to precisely and efficiently manipulate objects of all shapes in any environment, but until recently, not many works have been done to fully understand haptic feedback. This work proposed a novel method for getting a better shape descriptor than existing methods for classifying an object from multiple tactile data collected from a tactile glove. It focuses on improving previous works on object classification using tactile data. The major problem for object classification from multiple tactile data is to find a good way to aggregate features extracted from multiple tactile images. We propose a novel method, dubbed as Tactile-ViewGCN, that hierarchically aggregate tactile features considering relations among different features by using Graph Convolutional Network. Our model outperforms previous methods on the STAG dataset with an accuracy of 81.82\%.
\end{abstract}


%
\IEEEpeerreviewmaketitle

\section{\textbf{Introduction}}
For humans, tactile information is essential to perform day to day activities. We can get information like shape, weight, inertial property of an object with just the interaction of our hand and an object, due to our ``sense of touch". In an experiment conducted by Westling et al. \cite{1}, they injected anaesthesia in the hands of volunteers and noticed that for humans, without tactile feedback, their movement while grasping objects became inaccurate and unstable. This experiment shows that humans rely on vision along with tactile feedback to perform everyday manipulation tasks. This could also be true for robots, so robotic manipulation could be made much better if we include haptics information along with a vision. Understanding and modelling this tactile grasp information could improve AR/VR experience and robotic object manipulation. But to get such information from tactile data is difficult. There has not been much work focused on that. Being able to find the inertial properties of objects from tactile data will be helpful, especially for tasks such as robotic manipulation of an object with a complex shape. \\
\vspace{0.5 mm}

In this work, we only focus on object classification using tactile data by getting a shape descriptor from multiple tactile images. One of the main problem is to aggregate features from multiple tactile data to get a shape descriptor. Previous work of Sundaram et al. \cite{2} used to solve this problem by concatenating or pooling multiple features to get a single feature. But, in this method, any permutation of the tactile features still represents the same object. Also, this approach doesn't consider any relations between the feature vectors. Some other approaches view this as a sequential problem and use CNN-LSTM \cite{3} or ConvLSTM \cite{4}. In our work, we propose a novel method called Tactile-ViewGCN which is based on View-GCN \cite{8} that considers the relationship between multiple tactile images. After evaluating our proposed method on STAG dataset [2], our model outperforms previous methods on the same dataset with an accuracy of 81.82\%.

\section{\textbf{Related Work}}
\subsection{Learning human grasp from tactile data}
A deep convolutional model is used in the work done by Sundaram et al. \cite{2} to get image features. Specifically, a modified ResNet-18 \cite{16} is used to get features from tactile images. To better classify the objects, they increased the number of input images used and get better results, which seems sensible since having more data about the object's contact could give a better idea about object shape. However, when multiple tactile images are used to classify the objects, finding an appropriate way to aggregate features from all tactile images will help to get a better global shape descriptor feature that could be used for classification. In \cite{2}, they have concatenated tactile features from multiple inputs to obtain a global feature that was used for object classification, weight detection, etc. But their approach doesn't consider the relation between multiple tactile images.  \\
\vspace{0.5 mm}

Other approaches consider this a sequential problem and try to predict object class using sequential models like ConvLSTM \cite{4}, CNN-LSTM \cite{3} variants, etc. However, these approaches didn't work well for this dataset. In work done by Wang et al. \cite{5}, they included IMU data along with tactile frame to include hand movement during object interaction. Still, no model configuration has provided any improvement over the model that uses only tactile data. They also didn't consider using multiple tactile frames to cover the whole surface of the object or the relation between multiple tactile frames. 
\begin{figure*}[t]
\includegraphics[scale = 0.5]{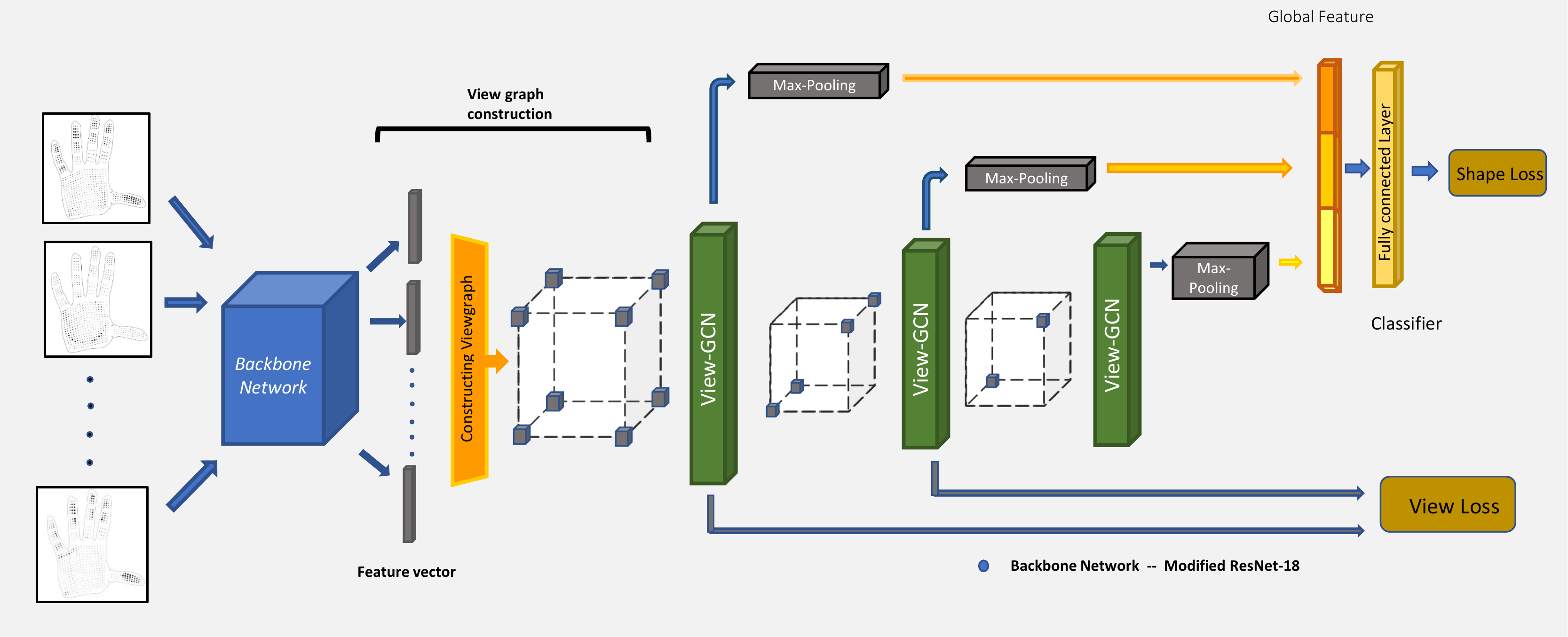}
\centering
\caption{Overview of our method, first a viewgraph is constructed from features from the Backbone CNN network (modified Resnet-18 for takking single channel input). Then, the viewgraph is hierarchically coarsened by multiple layers of view-GCN. Then, we concatenate features from all the levels to get a global shape descriptor which is passed to a classifier to predict the object class.}
\centering
\end{figure*}

\subsection{3-D shape recognition from multiple view images}
The view-based approach is advantageous in 3D shape recognition from multi-view images. Some methods like MVCNN \cite{6} get features from multiple images and aggregate them by taking max-pooling of all feature vectors to get a global shape descriptor. However, the relationship between features and discriminative properties of the views is left unexplored. Recent work \cite{7}, \cite{12} has shown that viewpoint location plays an essential role in getting a better shape descriptor for 3D shape recognition. Models like GVCNN \cite{11} has tried to include this by a hierarchical view-group-shape architecture of content descriptions. RotationNet \cite{7} considers a predefined viewpoint as a latent variable and tries to optimize it during training to find the best viewpoint for finding shape descriptors. 

In view-GCN \cite{8}, they use a viewgraph to represent the predefined viewpoints. The viewgraph representation enables them to use a modified GCN \cite{15} to aggregate features from multiple inputs, considering the relation between predefined viewpoints. Recent work done in GCN \cite{9} makes them perform spatial convolution by aggregating features in the local neighbourhood in the graph to update node features. So, in this work, we will be using the same method as view-GCN to aggregate multiple features from the tactile images with different backbone to see whether this method can give a better shape descriptor vector to classify objects from a tactile image.

\section{\textbf{Proposed Method}}
\subsection{Motivation}
For classifying an object using multiple tactile data from different regions of an object, no previous work has considered using the information about which part of the object the tactile image belongs to or the relationship between multiple tactile images, which can tell a lot about the shape of an object. Since some view-based approaches in 3D shape recognition have solved this problem by having a better way to aggregate features from multi-view images. We proposed to use a model called View-GCN, which has a good method for feature aggregation from multi-view images using Graph Convolutional Network.

\subsection{Model Architecture}
To get image features from tactile data, we use a Modified ResNet-18 which takes a single channel tactile image as input rather than a 3-channel RGB image and gives a one-dimensional feature vector as output. Then, we create a graph with nodes of the graph as a tactile image feature, obtained earlier for each input frame. A GCN is used to hierarchically combine features to get a shape descriptor of the object. View-GCN \cite{8} has three parts: 1) Local graph convolution, 2) Non-local message passing, and 3) Selective view sampling for coarsening the graph.

In part first, local graph convolution, the node features are updated by considering neighbouring nodes, which are found by applying kNN of viewpoints coordinates. Updated features from local graph convolution are passed through non-local message passing to include long range relationship between the graph's nodes. Then, selective view sampling is used to coarsen the graph, and they learn to choose the best view from the neighbouring view. \\
\vspace{0.5 mm}

\subsubsection{Building the view-graph}
View-Graph is built considering $i^{th}$-node of the graph as $i^{th}$ vertex of a cube, where the centre of the palm is assumed to be placed, while facing the centre of the cube when interacting with the object. The adjacent matrix of the graph is calculated as follow,
\begin{equation} \label{eq1}
S_{ij} = \Omega\bigl(g_{ij};\theta_{s}\bigr) 
\end{equation}
where, $g_{ij} = \bigl[v_i,v_j,v_{i} - v_{j},||v_{i} - v_{j}||^2\bigr] \in R^{10} $, and $g_{ij}$ represents the relation between two different viewpoints. For $\Omega$, we have used a three-layered MLP with LeakyReLU and has ten hidden units in the first two layers, and the output is a scalar $S_{ij}$.

We then use kNN to find the nearest neighbour for each viewpoints and only keep edges between nodes which are $n$ nearest neighbours of nodes, with edges being the distance between the viewpoints. Here, $n$ is chosen as a constant.\\
\vspace{0.5 mm}

\subsubsection{Local Graph Convolution}
Local graph convolution is defined as,
\begin{equation} \label{eq2}
F_l = \Psi(A^l F_{in}^l W^l; \theta^l_c ) , 
\end{equation}
where, $A^l$ is the learnable adjacent matrix of view graph at $l$th level ($G^l$). $F^l_n$ are feature vector in row, and $W^l$ is a linear transformation. $\Psi$ is a non-linear transformation consisting of linear and batch normalization layer, which is used to update the node features $F^l$.\\
\vspace{0.5 mm}

\subsubsection{Non-Local Message Passing}
To get pair wise relation between two nodes, we define,
\begin{equation} \label{eq3}
m^l_{ij} = \tau\bigl(\bigl[F^l_i , F^l_j \bigr]; \theta^l_m \bigr), i,j=1,2...N^l
\end{equation} 
where, $F^l_i$ is the feature vector of $i$th node and $\tau$ is a single-layered MLP with LeakyReLU to find the relation between pairwise nodes.
Then, all the nodes are updated by fusing the cumulated message with the feature vector $F^l$.
\begin{equation} \label{eq4}
   f^l_i = \Omega\bigl([f^l_i , r^l_i ]; \theta^l_f \bigr) 
\end{equation}
where, $r^l_i = \sum_{j=1}^{N_l} m^l_{ji} $ and $\Omega$ is a fusion function with parameter $\theta^l_f$. In the implementation, it is a single layer MLP with Batch normalization and LeakyReLU activation. The output is a fused feature vector that is updated considering the pairwise relation between nodes of the view graph.

\hspace{1 mm}
\subsubsection{Selective view Sampling}
Typically, Furthest Point Sampling (FPS) is used for coarsening the graph 
\cite{14}. FPS samples each view that is furthest from other previously sampled views based on viewpoint coordinates. But this sampling method doesn't assure that the sampled views are representative for the downstream classification tasks. Because of that, our model uses selective view sampling.

In this, we first select a subset of viewpoints using FPS. Then, we choose a sampled view by view selector among the kNN viewpoints of sampled viewpoints. The maximum output of view selector in the local neighbourhood is the new sampled view. The viewpoint coordinate of sampled view is given by,
\begin{equation} \label{eq5}
v^{l+1}_j = \underset{v_q \in N(v_j)}{\operatorname{argmax}}\bigl(\operatorname{max}(V(f^l_{v,q};\theta^{l,j}_v))\bigr)
\end{equation} 
where, $V()$ is the view selector which output the probability of a tactile data from sampled view belonging to $N_c$ object shape and max value of the vector is taken by using a max operator. Here, for the view selector, we have used a single layer MLP with $d$ hidden units. View selectors are trained to select discriminative views among neighbouring viewpoints. We derive a coarsened graph with nodes selected by selective view sampling using this method.

This View-GCN is applied hierarchically to the graph $G^l$ and output graph $G^{l+1}$ at each level, which is a coarsened graph with updated nodes with fewer nodes than the previous graph. At each level, max pooling is applied on the node features after being updated by local graph convolution to get a pooled descriptor. The final global shape descriptor is the concatenation of all the pooled features from each layer.

\subsubsection{Training loss}
While training, we consider two loss functions, \textit{i.e.}, shape loss and view loss. The total loss is the sum of both, computed as
\begin{multline} \label{eq6}
L  =  L_{shape}\bigl(Y_o,Y)\bigr) + \\
\sum_{l=0}^{L-1}\sum^{N_l +1}_{j=1}\sum_{v_q\in N(v_j)}L_{view}\bigl(V(f^l_{v_q};\theta^{l,j}_v),Y\bigr)
\end{multline} 
where, $L_{shape}$ is cross-entropy loss based on the output class we got and $L_{view}$ ensures that each view selector is able to classify object category based on view features from the local neighbourhood.

\begin{figure}[t]
\includegraphics[scale = 0.3]{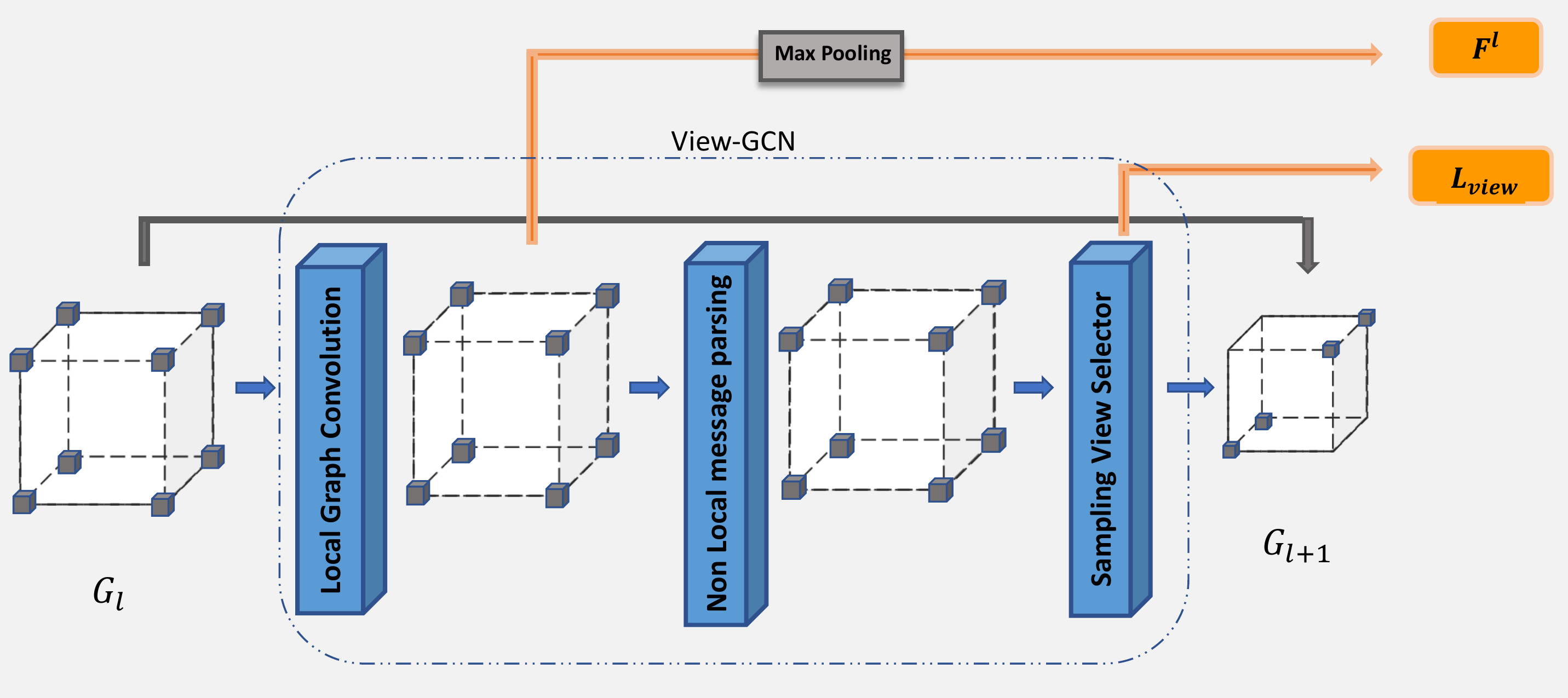}
\caption{View-GCN consists of 3 parts, local graph convolution, non-local message passing and selective view sampling for coarsening the view graph.}
\end{figure}

\section{\textbf{Experiment Analysis and Results}}
\subsection{Dataset}
For training this model, we used MIT-STAG dataset \cite{2}. This dataset is collected with a scalable tactile glove designed by Sundaram et al. \cite{2} covering the full hand with 1024 sensors. Using that, they created a large dataset with 135,000 frames, while interacting with 26 objects with a hand. After removing frames with no useful information, we have a training dataset of 36,531 frames and a testing dataset with 16,119 frames. The tactile image in the dataset is of size $32\times$32 with a single channel. To ensure that, these tactile signals correspond only to the interactions between the hand and objects, measured tactile signals are subtracted from the tactile signals obtained from an empty hand.

\begin{figure*}[t]
\centering
\includegraphics[scale = 0.50]{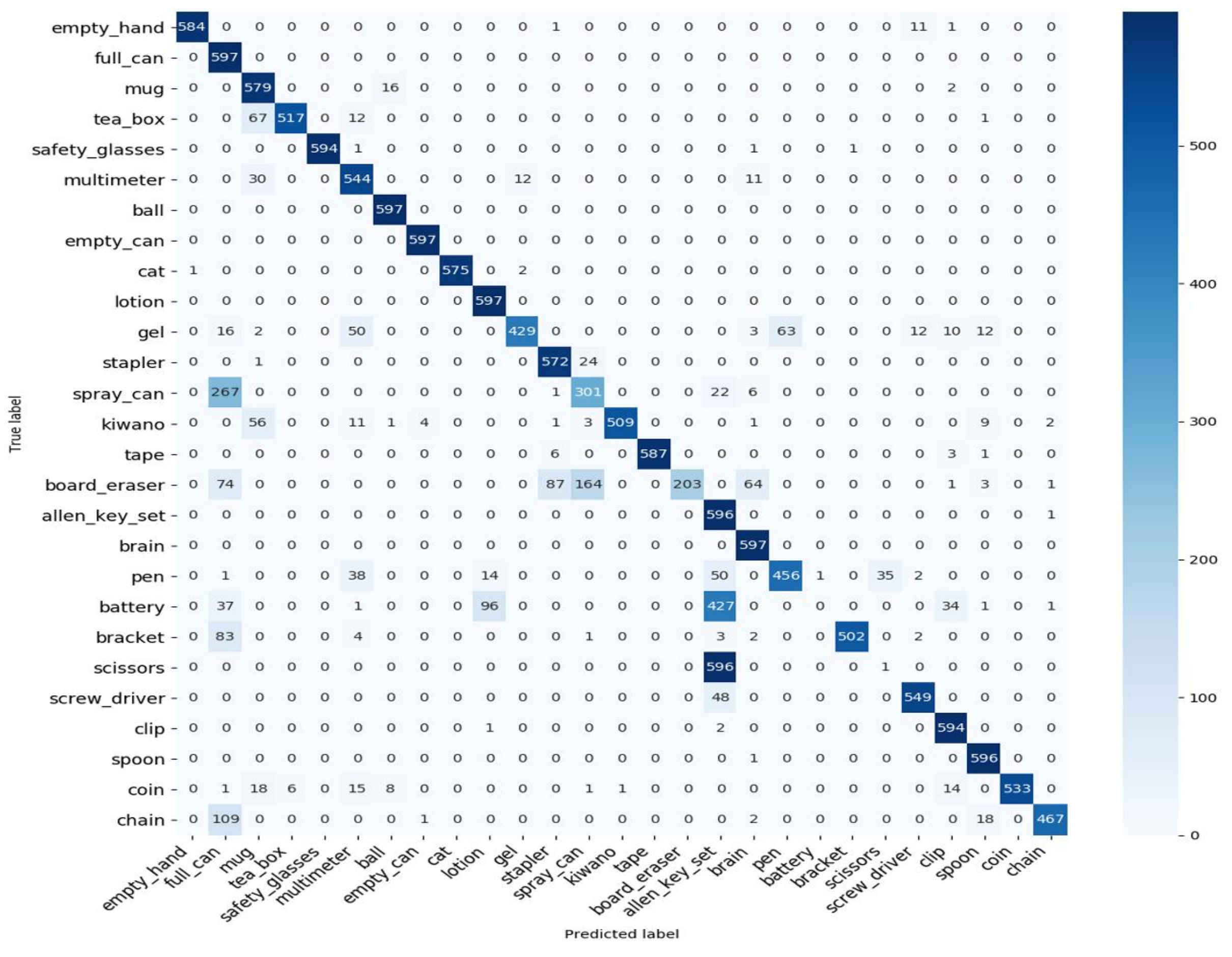}
\caption{Confusion matrix of our model which is trained on clustered dataset with 8 predefined views.}
\label{confusion_matrix} 
\end{figure*}

\subsection{Implementation Detail}
Tactile-viewGCN is trained considering that the tactile data are collected from predefined viewpoints. Since dataset doesn't have IMU data for respective frames, we won't be able to calculate the viewpoints for the tactile image. So, we have used kNN to cluster each object type in the dataset into 8 clusters, and each cluster is assumed to be data collected from viewpoints places at vertices of a cube.

Training consists of two parts. First, a convolutional backbone (a modified Resnet-18) is used to learn features of the tactile images by training it from scratch to classify tactile image datasets without clustering the dataset. It is trained for 30 epoch with optimized learning rate, weight decay, and momentum as \num{5e-3}, \num{1e-4}, 0.9, respectively. The learning rate is scheduled to decay by half for every ten epoch.

Then, the whole model is trained for 15 epoch with a learning rate of \num{1e-4}, \num{5e-4} for backbone and view GCN, respectively, with weight decay and momentum the same as before and everything is implemented using PyTorch \cite{13}. Since the View-GCN requires predefined viewpoints, we use a clustered dataset for training. This model is trained on Nvidia GeForce GTX 1650 GPU. The model is tested on a dataset [2] which is clustered using kNN methods into $N$ clusters, where $N$ is dependent on the number of input images used.

\subsection{Comparative Analysis}
The accuracy for our model fluctuated around 82\%, which is expected since we are clustering the dataset and the test dataset distribution is different each time. It can also be due to our assumption of the predefined viewpoint position. So, we have taken an average of 10 trials to calculate the model's accuracy. We can see that using View-GCN to aggregate features from multiple tactile images gives better results than just concatenating features to get a shape descriptor as in \cite{2}, \cite{5}. Our method has given 9.44\% more accuracy than the previous method used.

\begin{table}[ht]
\centering
\caption{Comparison on Stag \cite{2} Dataset}
\begin{tabular}{|c | c | c |} 
 \hline
 Method & Number of Input Image & Accuracy \\ [0.5ex] 
 \hline
 Sundaram et al \cite{2} & 7 & 72.38 \\ 

 Wang et al \cite{5} & 8 & ~72  \\

 Ours & 8 & 81.82  \\
 \hline
\end{tabular}

\end{table}%

\subsubsection{Effect of changing predefined viewpoints}
First, we chose the predefined viewpoints as vertices of a cube. We also tried a 12 view circular configuration with an elevation of 30 degrees around the upright direction.

\begin{table}[ht]
\centering
\caption{Comparison of model performance with different viewpoint configuration}
\begin{tabular}{|c | c | c |} 
 \hline
 Number of views & Configuration of viewpoints & Accuracy \\ [0.5ex] 
 \hline
 8 & cubic & 81.82  \\

 12 & circular & 80.26  \\
 \hline
\end{tabular}

\end{table}

\subsubsection{Effect of not using clustered Dataset}
When we trained the model on an unclustered dataset, the classification accuracy we got was 76.19\%, which is 5.63\% less than what we got using a clustered dataset. This decrease in accuracy could be due to the high correlation among input images since they are sampled from an unclustered dataset. Thus, the dataset cannot mimic the assumption that the collected images are from predefined viewpoints.

\subsubsection{Failure cases}
As we can see from the confusion matrix of our model from Fig.~\ref{confusion_matrix}, our model performed well on all classes except scissors, battery and board eraser. Scissors and the battery have been mostly misclassified as Allen key set, which has decreased the overall performance of the model.

\section{\textbf{Conclusion}}
This paper assumed that object recognition from tactile data problems could be considered similar to 3D shape recognition from multiple view images. Our results show that finding a better method to aggregate features from multiple tactile images considering the relation between the features performs better than the previous methods. 

The proposed model performance could be improved if the dataset contained any IMU data so that we might be able to calculate predefined viewpoints for each data, and thus eliminating the assumption that clustering the dataset might mimic data collected from predefined viewpoints.

\bibliographystyle{IEEEtran}
\bibliography{bare_conf}
%

\end{document}